\documentclass{article}

\usepackage[final]{neurips_2022}

\usepackage[utf8]{inputenc} 
\usepackage[T1]{fontenc}    
\usepackage{hyperref}       
\usepackage{url}            
\usepackage{booktabs}       
\usepackage{amsfonts}       
\usepackage{nicefrac}       
\usepackage{microtype}      
\usepackage{xcolor}         

\usepackage{microtype}
\usepackage{graphicx}
\usepackage{subfigure}
\usepackage{booktabs} 

\usepackage{color}
\usepackage{mathtools}

\usepackage{amsmath,amsfonts,bm}

\def\eqref#1{equation~\ref{#1}}

\def\1{\bm{1}}

\def\va{{\bm{a}}}

\def\vr{{\bm{r}}}
\def\vs{{\bm{s}}}

\def\mA{{\bm{A}}}

\def\mD{{\bm{D}}}

\def\mM{{\bm{M}}}

\def\mR{{\bm{R}}}
\def\mS{{\bm{S}}}
\def\mT{{\bm{T}}}

\DeclareMathAlphabet{\mathsfit}{\encodingdefault}{\sfdefault}{m}{sl}
\SetMathAlphabet{\mathsfit}{bold}{\encodingdefault}{\sfdefault}{bx}{n}

\newcommand{\R}{\mathbb{R}}

\DeclareMathOperator*{\argmax}{arg\,max}

\usepackage{subfigure}
\usepackage{multirow}
\usepackage{makecell}
\usepackage{array, booktabs, arydshln, xcolor}
\usepackage{amssymb}
\usepackage{graphbox}
\usepackage{url}
\usepackage{algorithm} 
\usepackage{amsthm}
\usepackage{algpseudocode}

\usepackage{amsthm}

\usepackage{thmtools}
\usepackage{thm-restate}
\usepackage{hyperref}
\usepackage{cleveref}

\allowdisplaybreaks

\usepackage{color}
\usepackage{mathtools}

\usepackage{wrapfig, blindtext}
\usepackage{subfigure}
\usepackage{multirow}
\usepackage{makecell}
\usepackage{array, booktabs, arydshln, xcolor}
\usepackage{amssymb}
\usepackage{graphbox}
\usepackage{caption}
\allowdisplaybreaks

\newcommand{\shortn}{\textup{\texttt{-}}}

\newcommand{\ie}{\textit{i}.\textit{e}.}

\newcommand{\name}{\textsc{Solar}}
\newcommand{\Amorpheus}{\textsc{Amorpheus}}
\newcommand{\SMP}{{SMP}}
\newcommand{\Amorpheuswmask}{\Amorpheus{} \textit{w/ synergy mask}}
\newcommand{\namewopreference}{\name{} \textit{w/o preference}}
\newcommand{\MetaMorph}{MetaMorph}
\newcommand{\Monolithic}{Monolithic}
\newcommand{\Walkers}{$\mathtt{Walker}$++}
\newcommand{\Humanoids}{$\mathtt{Humanoid}$++}
\newcommand{\Hoppers}{$\mathtt{Hopper}$++}
\newcommand{\WHHs}{$\mathtt{Walker}$-$\mathtt{Humanoid}$-$\mathtt{Hopper}$++}
\newcommand{\UNIMALS}{UNIMALS}
\newcommand{\Locomotion}{Locomotion}
\newcommand{\Manipulation}{Manipulation}

\newcolumntype{L}{>{$}l<{$}}
\newcolumntype{C}{>{$}c<{$}}
\newcolumntype{R}{>{$}r<{$}}
\newcolumntype{P}[1]{>{\centering\arraybackslash}p{#1}}

\title{Low-Rank Modular Reinforcement Learning via Muscle Synergy}

\author{%
  Heng Dong$^*$ \\
  IIIS, Tsinghua University\\
  \texttt{drdhxi@gmail.com} \\
  \And
  Tonghan Wang$^*$ \\
  Harvard University \\
  \texttt{twang1@g.harvard.edu} \\
   \AND
   Jiayuan Liu \\
   IIIS, Tsinghua University \\
   \texttt{georgejiayuan@gmail.com} \\
   \And
   Chongjie Zhang \\
   IIIS, Tsinghua University \\
   \texttt{chongjie@tsinghua.edu.cn} \\
}

\begin{document}

\maketitle

\def\thefootnote{*}\footnotetext{These authors contributed equally to this work.}

\begin{abstract}
Modular Reinforcement Learning (RL) decentralizes the control of multi-joint robots by learning policies for each actuator. Previous work on modular RL has proven its ability to control morphologically different agents with a shared actuator policy. However, with the increase in the Degree of Freedom (DoF) of robots, training a morphology-generalizable modular controller becomes exponentially difficult. Motivated by the way the human central nervous system controls numerous muscles, we propose a Synergy-Oriented LeARning (\name{}) framework that exploits the redundant nature of DoF in robot control. Actuators are grouped into synergies by an unsupervised learning method, and a synergy action is learned to control multiple actuators in synchrony. In this way, we achieve a low-rank control at the synergy level. We extensively evaluate our method on a variety of robot morphologies, and the results show its superior efficiency and generalizability, especially on robots with a large DoF like \Humanoids{} and UNIMALs. 
\end{abstract}

\section{Introduction}

Deep reinforcement learning (RL) has contributed significantly to the sensorimotor control of both simulated~\citep{heess2017emergence,zhu2020robosuite} and real-world~\citep{levine2016end,mahmood2018benchmarking} robots. Monolithic learning is a popular paradigm for learning control policies. In this paradigm, a policy inferring a joint action for all limb actuators based on a global sensory state is learned. Although monolithic learning has made impressive progress~\citep{chen2020randomized,kuznetsov2020controlling}, it has two major shortcomings. First, the input and output space is large. For robots with more joints, learning controlling policies puts a heavy burden on the representational capacity of neural networks. Second, the input and output dimensions are fixed, making it inflexible to transfer the learned control policies to robots with different morphologies. 

Modular reinforcement learning provides an elegant solution to these problems. In this learning paradigm~\citep{wang2018nervenet}, the control policy is decentralized~\citep{peng2021facmac}, and each limb actuator is controlled by a local policy. Recent research efforts show that the local policies can learn high-performance and transferable control strategies by sharing parameters~\citep{huang2020one}, communicating to each other by message passing~\citep{huang2020one}, and adaptively paying attention to other actuators via graph neural networks~\citep{kurin2020cage}. By exploiting the flexibility and generalizability provided by modularity, a modular policy can now control robots of up to thousands of morphologies~\citep{gupta2021metamorph}.

Despite the significant progress, modular reinforcement learning is still limited in terms of the complexity of morphological structures that can be controlled and struggles on robots with many joints like Humanoid~\citep{kurin2020cage}. The large degree of control freedom presents a major challenge for learning control policies. A question is why humans can control hundreds of muscles with dexterity while the most advanced RL policy can only control less than fifteen actuators.

Studies on muscle synergies~\citep{d2003combinations} may provide an answer. A human central nervous system decreases the control complexity by producing a small number of electrical signals and activating muscles in groups~\citep{ting2007neuromechanics}. Muscle synergy is the coordination of muscles that are activated in synchrony. With muscle synergies, the human nervous system achieves low-rank control over its actuators. In this paper, we aim to use the inspiration of muscle synergies to reduce the control complexity and improve the learning performance of modular RL. 

The first challenge of incorporating muscle synergies into modular RL is to discover a synergy structure that can promote policy learning. Neuroscience researchers factorize electrical signals~\citep{saito2018muscle,falaki2017motor,kieliba2018muscle} to analyze the synergy structure, but policy signals are sub-optimal or even absent during reinforcement learning. We thus exploit the functional similarity and morphological context of actuators and use a clustering algorithm to identify actuators in the same synergy. The intuition is that muscles in a synergy typically serve the same functional purpose and have similar morphological contexts. We quantify the functional similarity by the influence of an actuator's action on the global value function, and the morphological structure is encoded as a distance matrix. To use the two types of information simultaneously, we adopt the affinity propagation algorithm~\citep{frey2007clustering}. The synergy structure is updated periodically during learning to promptly reflect changes in value functions.

To exploit the discovered synergy structure, we design a synergy-aware architecture for policy learning. The major novelty here is that the policy learns action selection for each synergy, and the synergy actions are transformed linearly to get actuator actions. Since the number of synergies is typically much smaller than actuators, we actually learn a low-rank control policy where the physical actions are a linear mapping from a low-dimensional action space. Moreover, for better processing state information, the synergy-aware policy adopts a two-level transformer structure, which first aggregates information within each synergy and then processes information across synergies.

We evaluate our Synergy-Oriented LeARning (\name{}) framework on two MuJoCo~\citep{todorov2012mujoco} locomotion benchmarks~\citep{huang2020one,gupta2021embodied} and in multi-task to zero-shot learning, single-task settings. \name{} significantly outperforms previous state-of-the-art algorithms in terms of both sample efficiency and final performance on all tested settings, especially on robots with a large DoF like \Humanoids{}~\citep{huang2020one} and UNIMALs~\citep{gupta2021embodied}. Performance comparison and the visualization of learned synergy structures strongly support the effectiveness of our synergy discovery method and synergy-aware transformer-based policy learning approach. Our experimental results reveal the \emph{low-rank} nature of multi-joint robot control signals.
\section{Background}\label{sec:related_work}

\textbf{Modular RL}. Modular Reinforcement Learning decentralizes the control of multi-joint robots by learning policies for each actuator. Each joint has its controlling policy and they coordinate with each other via various message passing schemes. Modular RL usually needs to deal with agents with different morphologies. To do so, \citet{wang2018nervenet} and \citet{pathak2019learning} represent the robot's morphology as a graph and use GNNs as policy and message passing networks. \citet{huang2020one} uses both bottom-up and top-down message passing scheme through the links between joints for coordinating. All of these GNN-like works show the benefits of modular policies over a monolithic policy in tasks tackling different morphologies. However, recently, \citet{kurin2020cage} validated a hypothesis that any benefit GNNs can extract from morphological structures is outweighed by the difficulty of message passing across multiple hops. They further propose a transformer-based method, \Amorpheus{}, that utilizes self-attention mechanisms as a message passing approach. \Amorpheus{} outperforms prior works and our work is based on \Amorpheus{}. Previous works mainly focused on effective message passing schemes, while our work aims at reducing learning complexities when the DoF of the robot is large.
 
\textbf{Muscle Synergy}. How the human central nervous system (CNS) coordinates the activation of a large number of muscles during movement is still an open question. According to numerous studies, the CNS activates muscles in groups to decrease the complexity required to control each individual muscle~\citep{d2003combinations,ting2007neuromechanics}. According to muscle synergy theory, the CNS produces a small number of signals. The combinations of these signals are distributed to the muscles~\citep{wojtara2014muscle}. Muscle synergy is the term for the coordination of muscles that activate at the same time~\citep{ferrante2016personalized}. A synergy can include multiple muscles, and a muscle can belong to multiple synergies. Synergies produce complicated activation patterns for a set of muscles during the performance of a task, which is commonly measured using electromyography (EMG)~\citep{tresch2002coordination,singh2018systematic}. EMG signals are typically recorded as a matrix with a column for activation signals for a moment and a row for activation of a muscle~\citep{rabbi2020non}. Factorisation methods on the matrix are used to extract muscle synergies from muscle activation patterns. Four most commonly
used factorization methods are non-negative matrix factorisation~\citep{steele2015muscle,schwartz2016dynamic,lee1999learning,rozumalski2017muscle,shuman2016repeatability,saito2018muscle} , principal component analysis~\citep{ting2005limited,ting2015neuromechanical,danion2010motor,falaki2017motor}, independent component analysis~\citep{hyvarinen2000independent,hart2013distinguishing}, and factor analysis~\citep{kieliba2018muscle,saito2015coordination}.

In the field of robot control, only a few works~\citep{palli2014experimental, wimbock2011synergy, ficuciello2016synergy} have exploited the idea of muscle synergy for dimensionality reduction to simplify the control. However, these works usually first use motion dataset from humans to obtain the synergy space and then learn to control in this synergy space. In contrast, our work learns the synergy space simultaneously with the control policy in the synergy space.

\textbf{Affinity propagation}~\citep{frey2007clustering} is a clustering algorithm based on multi-round message passing between input data points. It does not need to pre-define the number of clusters and proceeds by finding each instance an exemplar. Data points that choose the same exemplar belongs to the same cluster. 

Suppose $\{x_i\}_{i=1}^n$ is a set of data points. Define $\mS\in\mathbb{R}^{n\times n}$ as a similarity matrix. When $i\ne j$, the element $s_{i,j}$ at $i$th row and $j$th column is the similarity between $x_i$ and $x_j$, which can be measured as, for example, the negative squared distance of two data points. When $i=j$, the element $s_{i,j}$ represents how likely the corresponding instance is to become an exemplar. The vector of diagonal elements, $(s_{11}, s_{22}, \dots, s_{nn})$, is called \emph{preference}. Non-diagonal elements in $\mS$ constitute the \emph{affinity} matrix. The algorithm takes $\mS$ as input and proceeds by updating two matrices: the responsibility matrix $\mR$ whose values $r_{i,j}$ represent whether $x_j$ is well-suited to be the exemplar for $x_i$; the availability matrix $\mA$ whose values $a_{i, j}$ quantify the appropriateness for $x_i$ picking $x_j$ as its exemplar~\citep{frey2007clustering}. These two matrices are initialized to be zeroes and can be regarded as log-probability tables. The algorithm then alternatives between two message-passing steps. First, the responsibility matrix is updated: 
\begin{equation}
r_{i,j} \leftarrow s_{i,j} - \max_{j' \neq j} \left( a_{i,j'} + s_{i,j'} \right).
\end{equation}
Then, the availability matrix is updated:
\begin{equation}
\begin{aligned}
     a_{i,j} &\leftarrow \min \Big( 0, r_{j,j} + \sum_{i' \not\in \{i,j\}} \max(0, r_{i',j}) \Big)\ \text{for } i\neq j; \quad
     a_{j,j} &\leftarrow \sum_{i' \neq j} \max(0, r_{i',j}).
\end{aligned}
\end{equation} 
Messages are passed until the clusters stabilize or the pre-determined number of iterations is reached. Then the exemplar of $i$ is $\argmax_{j} r_{i,j}+a_{i,j}$. 


\section{Method}\label{sec:method}

In this section, we present our Synergy-Oriented LeARning (\name) scheme that incorporates the muscle synergy mechanism into modular reinforcement learning to reduce its learning complexity. 

Our method has two major components. The first one is an unsupervised learning module that utilizes the morphological structure and value information to discover the synergy hierarchy. The second is a novel attention-based policy architecture that supports synergy-aware learning. Both of the components are specially designed to enable the control of robots with different morphologies. We first introduce the problem settings and then describe the details of the two components.

\textbf{Problem settings.} We consider $N$ robots, each with a unique morphology. Agent $n$ contains $K_n$ limb actuators that are connected together to constitute its overall morphological structure. Examples of such robots that are studied in this paper include \Humanoids{} and UNIMALs. At each discrete timestep $t$, actuator $k\in\{1,2,\dots,K_n\}$ of a robot $n\in \{1,2,\dots,N\}$ receives a local sensory state $s_t^{n,k}$ as input and outputs individual torque values $a_t^{n,k}$ for the corresponding actuator. Then the robot $n$ executes the joint action $\va_t^n=\{a_t^{n,k}\}_{k=1}^{K_n}$ at time $t$, after which the environment returns the next state $\vs_{t+1}^n=\{s_{t+1}^{n,k}\}_{k=1}^{K_n}$ corresponding to all limbs of the agent $n$ and a collective reward for the whole morphology $r_t^n(\vs_t^n,\va_t^n)$. We learn a policy $\pi_\theta$ to generate actions based on states. The learning objective of the policy is to maximize the expected return on all the tasks:
\begin{equation}
    \mathcal{J}(\theta)=\mathbb{E}_\theta \sum_{n=1}^{N}\sum_{t=0}^\infty\left[\gamma^t r_t^{n}(\vs_t^n,\va_t^n)\right],
\end{equation}
where $\gamma$ is a discount factor. We adopt an actor-critic framework for policy learning. The critic is shared among all tasks and estimates the Q-function for each robot $n$:
\begin{equation}
    Q^{\pi_\theta}(\vs^n,\va^n)=\mathbb{E}_\theta \sum_{t=0}^\infty\left[\gamma^t r_t^n(\vs_t^n,\va_t^n)|\vs^n_0=\vs^n,\va^n_0=\va^n\right].
\end{equation}

\begin{figure}[htbp]
\centering
\includegraphics[width=\linewidth]{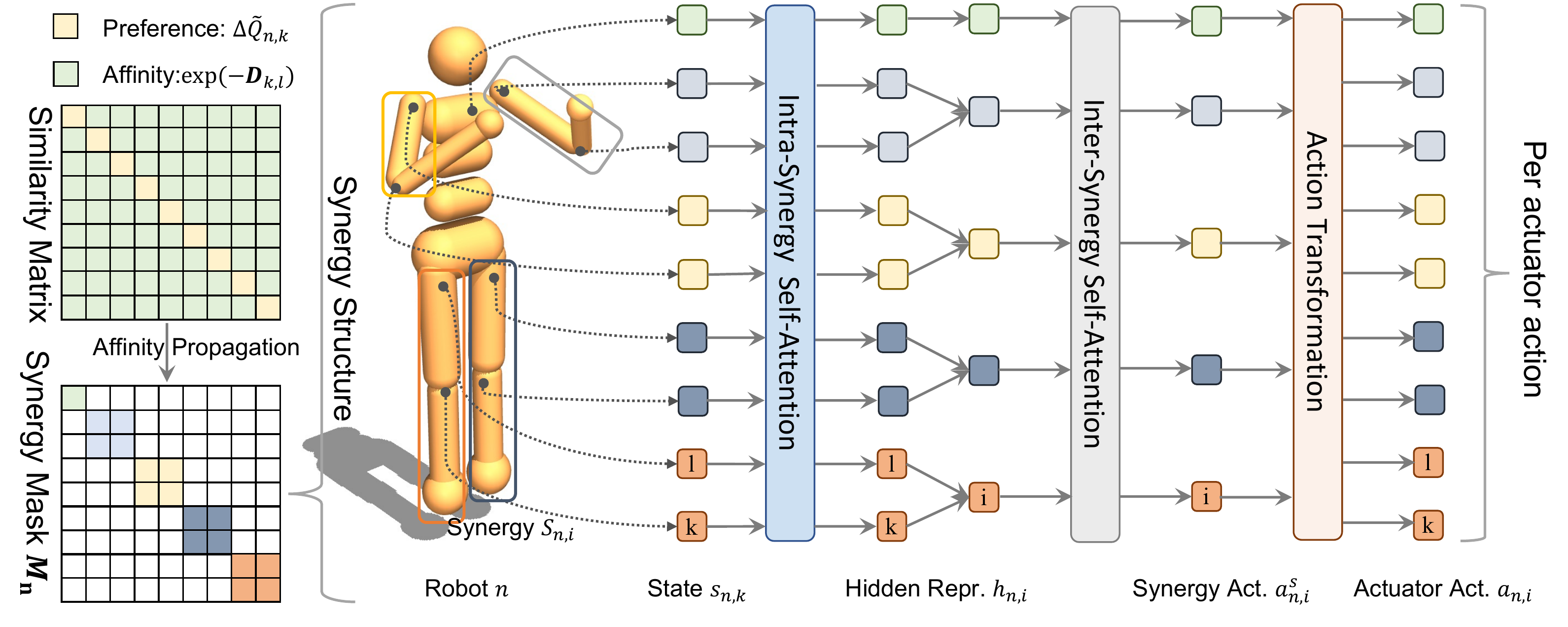}
\caption{Synergy-aware policy learning. The intra-synergy attention module aggregates actuator information within each synergy. The inter-synergy attention module synthesizes information from all synergies to produce synergy actions. Synergy actions are then transformed linearly to obtain actuator actions. Actuator actions are of a lower rank, reducing the control complexity. }\label{fig:framework}
\end{figure}

\subsection{Discovering synergy structure}\label{sec:affinity_prop}

In Neuroscience, the muscle synergies are usually discovered by a factorization of electrical muscle signals during performing tasks~\citep{todorov2004analysis,rabbi2020non}. Factorization is a method in hindsight -- it statistically analyzes the optimal control policies of animals embodied in the electrical signals. By contrast, in reinforcement learning, we do not have the optimal control policies in advance. The synergy hierarchy learned from non-optimal policies is more likely incorrect, which will hamper policy learning. Therefore, we propose learning the synergy hierarchy by an unsupervised learning method that incorporates morphological information besides learning information. In this section, we describe our synergy hierarchy discovery methods.

Intuitively, actuators in the same synergy are activated simultaneously and they together finish a motion of an end effector. This gives us a hint that actuators with similar functions are supposed to be in a synergy. Formally, the function of an actuator (the $k$th actuator in robot $n$) can be modelled by its influence to the value function:
\begin{equation}
   \Delta Q_{n,k} = \mathbb{E}_{\vs^n,a^{n,k},\va^{n,\shortn k}}\left[Q^\pi(\vs^n,[a^{n,k},\va^{n,\shortn k}]) - Q^\pi(\vs^n,[b^{n,k},\va^{n,\shortn k}])\right], 
\end{equation}
where $[\cdot,\cdot]$ combines two terms, $a^{n,k}$ is the actual action of actuator $k$, $b^{n,k}$ is a default action of actuator $k$ ($b^{n,k}=0$ in MuJoCo environments) and $\va^{n,\shortn k}$ is the actions of actuators on robot $n$ except for the $k$th one. In practice, we use a SoftMax function to regularize $\Delta Q_{n,k}$: $\Delta \tilde{Q}_{n,k}=\exp(\Delta Q_{n,k})/\sum_j \exp(\Delta Q_{n,j})$.

Ideally, $\Delta \tilde{Q}_{n,k}$ contains actuator clustering information, but the Q-values may be inaccurate during learning. We propose to treat $(\Delta \tilde{Q}_{n,1},\Delta \tilde{Q}_{n,2},\cdots, \Delta \tilde{Q}_{n,K_n})$ as the preference vector of the affinity propagation clustering algorithm (Sec.~\ref{sec:related_work}), i.e, the diagonal elements of similarity matrix, and in the next paragraph we further encode stable morphological information in the affinity matrix of affinity propagation, i.e, the non-diagonal elements of similarity matrix. The advantage of this design is that (1) we can simultaneously consider functional similarity and morphological context when using clustering to discover synergies and (2) static morphological information may help stabilize the clustering results.

We now discuss how to encode morphological information in the affinity matrix. A robot morphology is usually organized in a tree structure with its torso as the root node~\citep{huang2020one}. We define the adjacency matrix $\mA_d\in\{0,1\}^{K_n\times K_n}$ to represent robot's morphology, where $A_{i,j}=1$ if actuators $i,j$ are directly linked. To measure the connectiveness of every actuator pair, we use Floyd-Warshall algorithm~\citep{cormen2022introduction} to compute the shortest distances between actuators based on $\mA_d$, resulting in a distance matrix $\mD\in\mathbb{N}^{K_n\times K_n}$. The matrix $\exp(-\mD)$ will serve as the affinity matrix $\mA$ of affinity propagation. After defining the preference and affinity for affinity propagation, the clustering algorithms run $n_{max}$ rounds of message passing to determine actuator clusters. We treat each cluster as a synergy.

In practice, we update the synergy structure periodically during the learning process. While $\mD$ remains unchanged, $\Delta \tilde{Q}_{n,k}$ changes as the Q-function is updated. Each time we run the affinity propagation with the new preference vector and then freeze the synergy structure for a while.


\subsection{Learning synergy-aware low-rank policies}\label{sec:synergy_policy}

We now describe how we design a synergy-aware actor-critic learning framework that exploits the synergy hierarchy discovered in Sec.~\ref{sec:affinity_prop}.

The core idea is that the policy generates control signals (actions) for synergies instead of actuators. As the number of synergies is much fewer than actuators, the learning complexity is reduced. The synergy actions will be combined linearly to generate physical actions for actuators. In this way, we learn a low-rank control policy -- the physical actions are actually a linear combination of actions in a low-dimensional space.

Suppose that we find $L_n$ synergies $\{S_{n,i}\}_{i=1}^{L_n}$ for the $n$th robot. We use a mask $\mM_n\in\{0,1\}^{K_n\times K_n}$ to represent the synergy structure. The element at the $i$th row, $j$th column of $\mM_n$ is 1 if actuator $i$ and $j$ are in the same synergy. The policy $\pi_\theta$ consists of three components: an intra-synergy attention module, an inter-synergy attention module, and an action transformation matrix as in Figure~\ref{fig:framework}. 

The first module aggregates information of actuators in a synergy. Input to this module is the joint state $s_n$, and this module outputs a hidden representation $h_{n,i}\in\mathbb{R}^d$ for each synergy $S_{n,i}$. In practice, we use a transformer with a two-head self-attention layer where the attentions between actuators belonging to different synergies are masked out. The output of the self-attention layer is of the size $K_n\times d$. We then use mean pooling to aggregate information of actuators in the same synergy. In this way, we get a hidden state $h_n\in \mathbb{R}^{L_n\times d}$, with each row being the aggregated information of a synergy.

The second module aggregates information from different synergies and outputs the synergy actions. It has two multi-head self-attention layers. We feed $h_n$ into this these layers and get vector-valued synergy actions $\va^s_n\in\mathbb{R}^{L_n\times 1}$. The synergy actions are then transformed by learnable matrices $\mT\in \mathbb{R}^{K_n\times L_n}$, and we obtain actions for actuators by $\va_n = \mT\va^s_n$. One question about the transformation matrix $\mT$ is that $L_n$ may change during the learning process, requiring the dimension of $\mT$ to be dynamic. In Appendix~\ref{appx:A}, we discuss how to deal with this problem.


The critic uses the same intra- and inter-synergy attention structure as the policy for processing state information. The outputs of the inter-synergy attention module are fed into a fully-connected network to obtain the Q-value estimation. In practice, we use TD3~\citep{fujimoto2018addressing} for the training of the policy and the critic. The intra- and inter-synergy transformer and the transformation matrix are updated in an end-to-end manner. We provide more details of our learning framework in Appendix~\ref{appx:A}.



\section{Experiments}
In this section, we benchmark our method \name{} on various MuJoCo~\citep{todorov2012mujoco} locomotion tasks. we evaluate the effectiveness of \name{} by asking the following questions: (1) Can our \name{} outperforms other modular RL approaches when simultaneously trained on a large number of diverse agents with different morphologies? (Sec.~\ref{sec:exp_multitask}) (2) Can the learned policy in multi-task environments generalize to new tasks with unseen morphologies? (Sec.~\ref{sec:exp_zeroshot}) (3) How does \name{} learn synergy clusters and how do synergy clusters facilitate learning? (Sec.~\ref{sec:exp_analysis}) (4) Can \name{} scale to single-tasks with numerous joints? (Sec.~\ref{sec:exp_singletask}). For qualitative results, please refer to the videos on our project website\footnote[1]{\url{https://sites.google.com/view/synergy-rl}}. And our code is available at GitHub\footnote[2]{\url{https://github.com/drdh/Synergy-RL}}.

\subsection{Experiment setup}\label{sec:exp_setup}
We run experiments on two benchmarks, and report several results in this section. For additional results, please refer to Appendix~\ref{appx:add_exp}. We test all methods with 4 random seeds and show the mean performance as well as 95\% confidence intervals.

\textbf{Environments}. For multi-task and zero-shot evaluation, we adopt the widely-used modular MTRL benchmarks \citep{huang2020one,kurin2020cage,hong2021structure}, which are created based on Gym MuJoCo locomotion tasks by~\citet{huang2020one}. In this section, we report results on 3 in-domain settings: (1) 6 variants of walker [$\mathtt{Walker}$++] (2) 8 variants of humanoid [$\mathtt{Humanoid}$++] (3) 3 variants of hopper [$\mathtt{Hopper}$++]; and 1 cross-domain settings: all 6 variants of walker, all 8 variants of humanoids, and all 3 variants of hopper [$\mathtt{Walker}$-$\mathtt{Humanoid}$-$\mathtt{Hopper}$++]. For single-task evaluation, we sample 3 morphologies from UNIMALS~\citep{gupta2021embodied, gupta2021metamorph}, which are obtained from evolution and have numerous joints, making it suitable to evaluate the performance of modular RL policies in single-tasks. For a detailed description of the environments please refer to~\citet{gupta2021embodied} and Appendix~\ref{appx:repro}.

\textbf{Baselines}. In this section, we compare our method \name{} against state-of-the-art modular RL methods \SMP{}~\citep{huang2020one} and \Amorpheus{}~\citep{kurin2020cage}. \SMP{} passes messages along the limbs using a bottom-up and top-down scheme. \name{} and \Amorpheus{} use transformer-based actors and critics, whose message passing schemes are realized by self-attention. We also compare \name{} with standard TD3-based non-modular RL: \Monolithic{}. Please refer to Appendix~\ref{appx:repro} for more details about baselines. 

\textbf{Ablations}. There are two contributions that characterize our method. (1) An unsupervised learning module that utilizes the value information and the morphological contextual structure to discover the synergy hierarchy. (2) A novel attention-based synergy-aware policy architecture that supports synergy-aware learning. Our novelties are mainly about the discovery and utilization of synergies, and these two contributions closely rely on each other. If we totally ablate synergies, we will get \Amorpheus{}. Therefore, the effectiveness of synergy-based learning can be demonstrated by comparing \name{} against \Amorpheus{}. We further design the following ablations: (1) \Amorpheuswmask{}. Incorporate synergy-aware inter-cluster masks into \Amorpheus{} to test the contribution of synergy hierarchy. (2) \namewopreference{}. Remove value information (the preference vector) from affinity propagation of \name{}. In this case, affinity propagation is based solely on morphological information, and we can test the effects of value information.

\textbf{Implementations}. We use TD3~\citep{fujimoto2018addressing} as the underlying reinforcement learning algorithm for training the policy over all baselines, ablations and our method for fairness. We implement \name{} in the \Amorpheus{} codebase. And like \Amorpheus{}, there is no weight sharing between actor and critic. \name{} uses a traversal-based embedding for action transformation, which may incorporate structural information. For a fair comparison, we concatenate these embeddings into the original observation vectors of each environment which is applied to all tested algorithms.

\subsection{Multi-task with different morphologies}\label{sec:exp_multitask}

\begin{figure}[htbp]
\centering
\includegraphics[width=\linewidth]{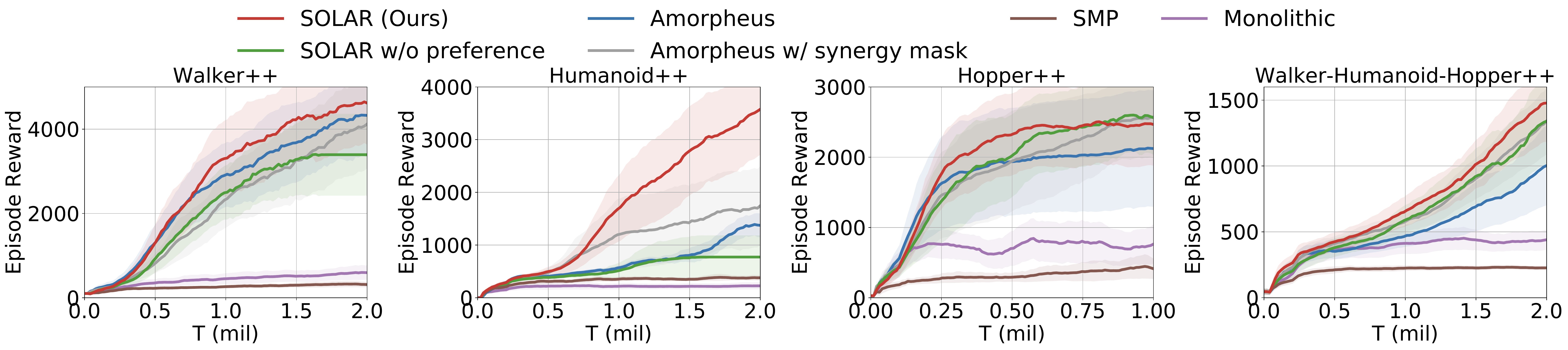}
\caption{Multi-task performance of our method \name{} compared to baselines and ablations.}\label{fig:MTRL}
\end{figure}

\begin{figure}[htbp]
\centering
\includegraphics[width=\linewidth]{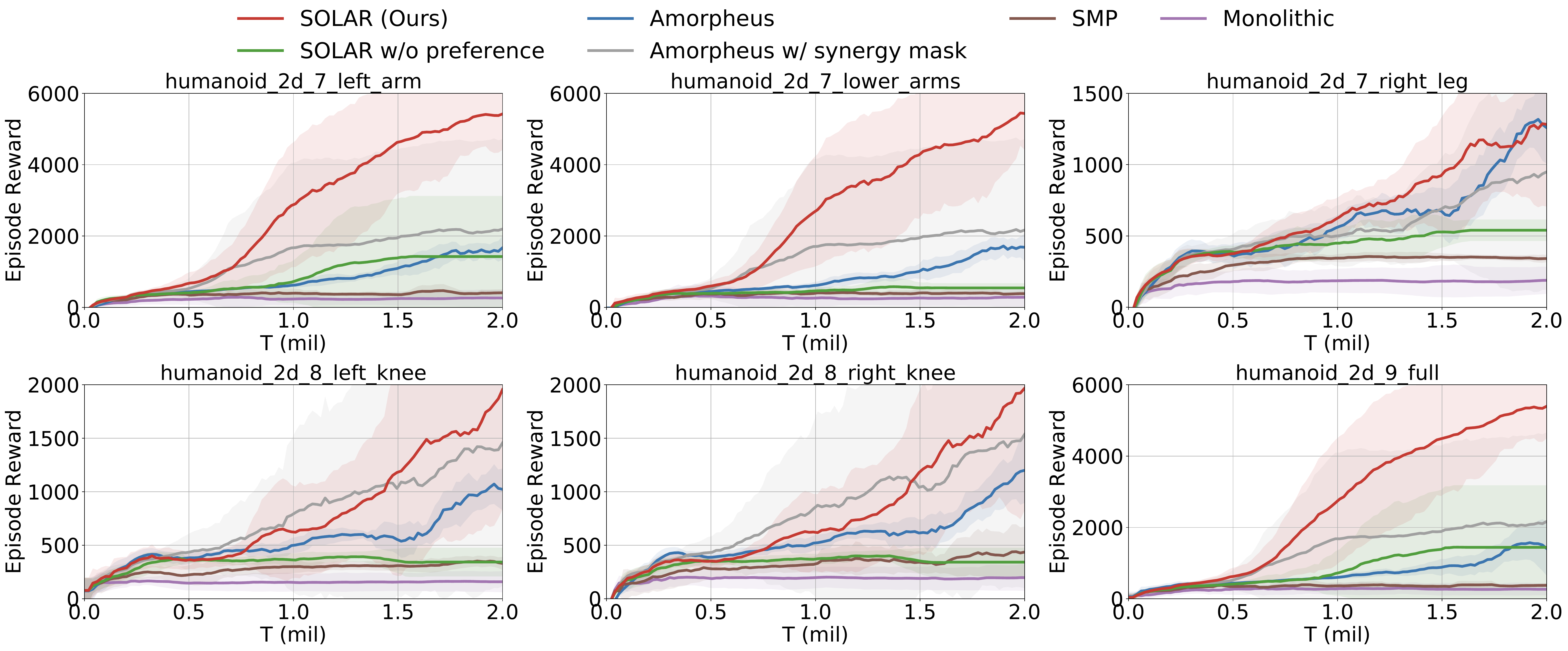}
\caption{Multi-task performance of our method \name{} compared to baseline and ablations on $\mathtt{Humanoid}$++.}\label{fig:MTRL_humanoids}
\vspace{-2em}
\end{figure}

We summarize the multi-task results in Figure~\ref{fig:MTRL}. \name{} outperforms the previous state-of-the-art algorithms \Amorpheus{} and \SMP{} in all settings. Furthermore, the performance gap between \name{} and \Amorpheus{} is notably larger in \Humanoids{} which has the maximum number of joints among the tested settings. We also plot the learning curves of six training variants (Figure~\ref{fig:MTRL_humanoids}) and two testing variants (Figure~\ref{fig:zeroshot}) of \Humanoids{}. These results consistently demonstrate the effectiveness of \name{} in training robots with a large degree of freedom (DoF). We speculate that the reason of the effectiveness is that \name{} controls robots in a lower-rank action space than the original space and the action spaces of robots with a large DoF are more likely to be reduced via the synergy mechanism. To further validate this speculation, we conduct extra evaluations on robots with much more joints in Sec.~\ref{sec:exp_singletask}. 

As for ablations, \Amorpheuswmask{} outperforms \Amorpheus{} in most tasks in Figure~\ref{fig:MTRL} and Figure~\ref{fig:MTRL_humanoids}, which indicates the effectiveness of the unsupervised learning method that discovers synergy clusters and synergy-aware masks in the policy. However, \namewopreference{} performs comparable to or even worse than \Amorpheus{}, which reflects the prominent usefulness of value information.

\subsection{Zero-shot generalization}\label{sec:exp_zeroshot}

\begin{figure}[htbp]
\centering
\includegraphics[width=\linewidth]{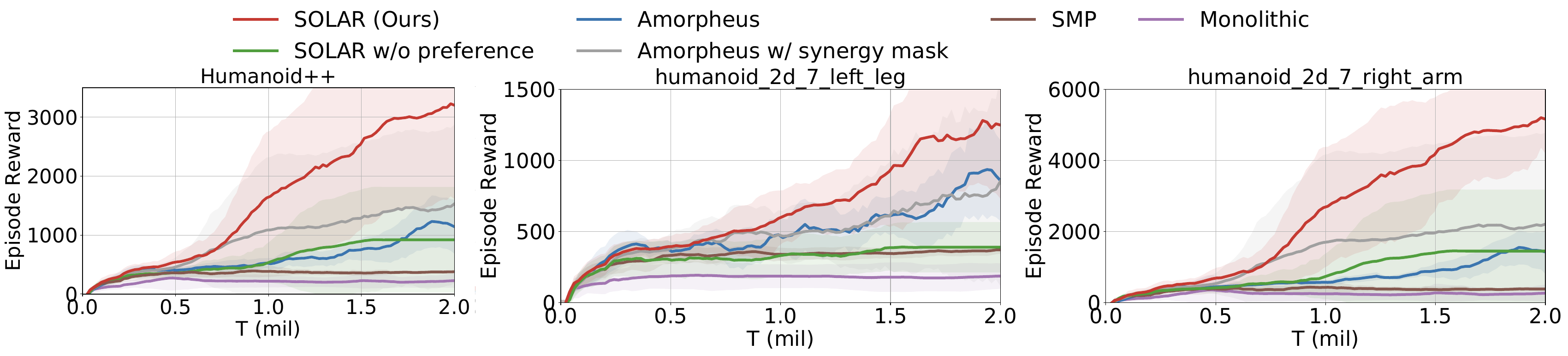}
\caption{Zero-shot performance of our method \name{} compared against baseline and ablations.}\label{fig:zeroshot}
\end{figure}

In this section, we benchmark \name{} in a zero-shot learning setting where the policy is trained in multiple tasks and then generalized to unseen tasks. We compare \name{} with \Amorpheus{}, \SMP{}, \Amorpheuswmask{} and \namewopreference{} in \Humanoids{}, which has the largest number of joints among the tested environments. \Humanoids{} has eight variants of humanoids, where six of them are used as training tasks (see Figure~\ref{fig:MTRL_humanoids}) and the other two are used as testing tasks (Figure~\ref{fig:zeroshot}). To obtain the results of testing tasks at each time step, we periodically load policy network parameters when training and evaluate the policy in these tasks. The trajectories on testing tasks will be discarded and will not be used for learning. 

As shown in Figure~\ref{fig:zeroshot}, \name{} outperforms all the baselines by a large margin in these unseen tasks. \name{} learns faster than other baselines and shows higher sample efficiency in Figure~\ref{fig:MTRL_humanoids}, showing that the knowledge obtained from training tasks is effectively used in unseen tasks.

\subsection{Analysis of synergies}\label{sec:exp_analysis}

\begin{figure}[htbp]
\centering
\includegraphics[width=\linewidth]{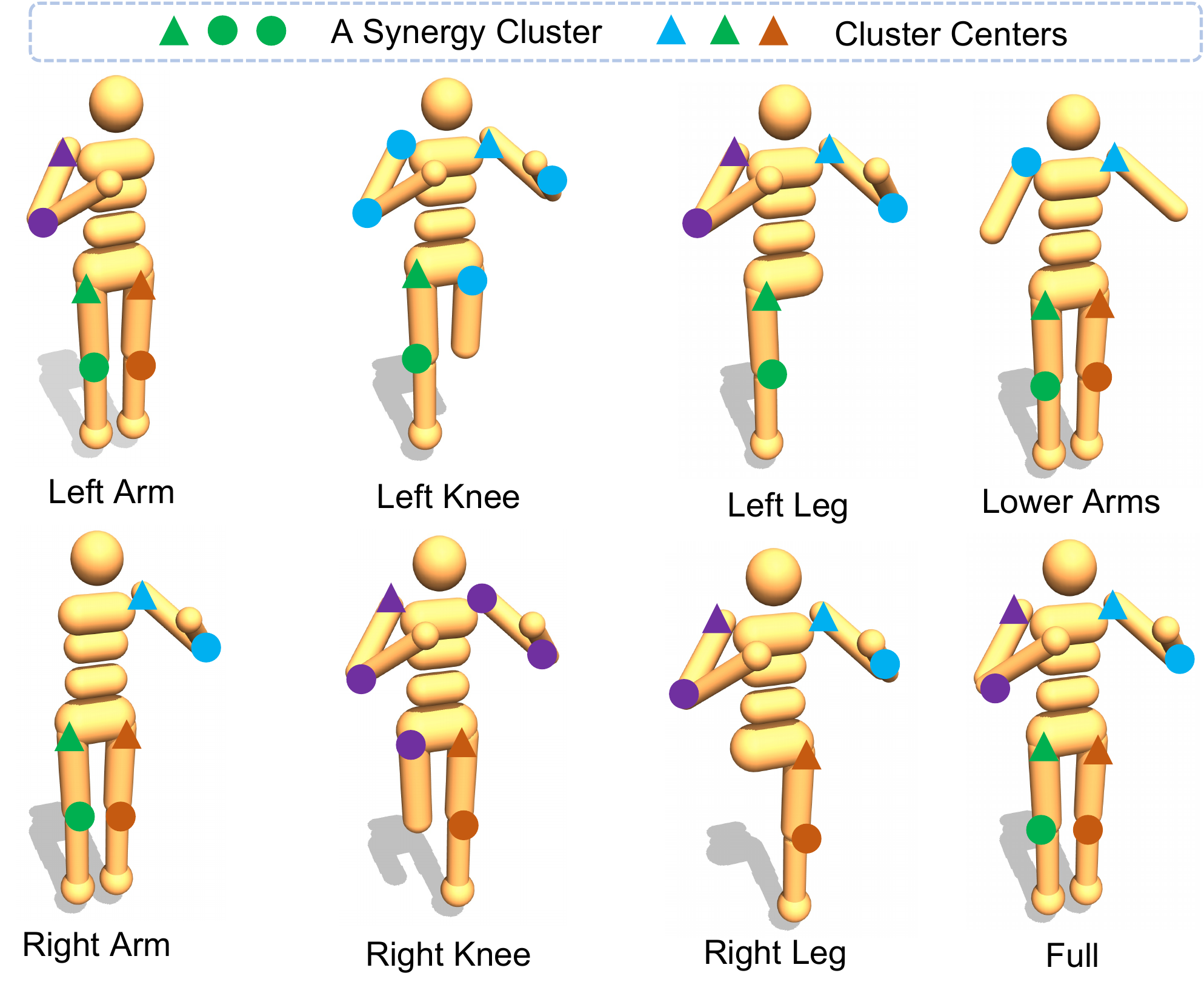}
\caption{Synergy clustering results of \name{} in \Humanoids{}. Different colors represent different synergy clusters, and joints marked with the same color are in the same cluster. Joints marked with triangles are the centers of their corresponding clusters.}\label{fig:synergy_cluster}
\end{figure}
\begin{figure}[htbp]
\centering
\includegraphics[width=\linewidth]{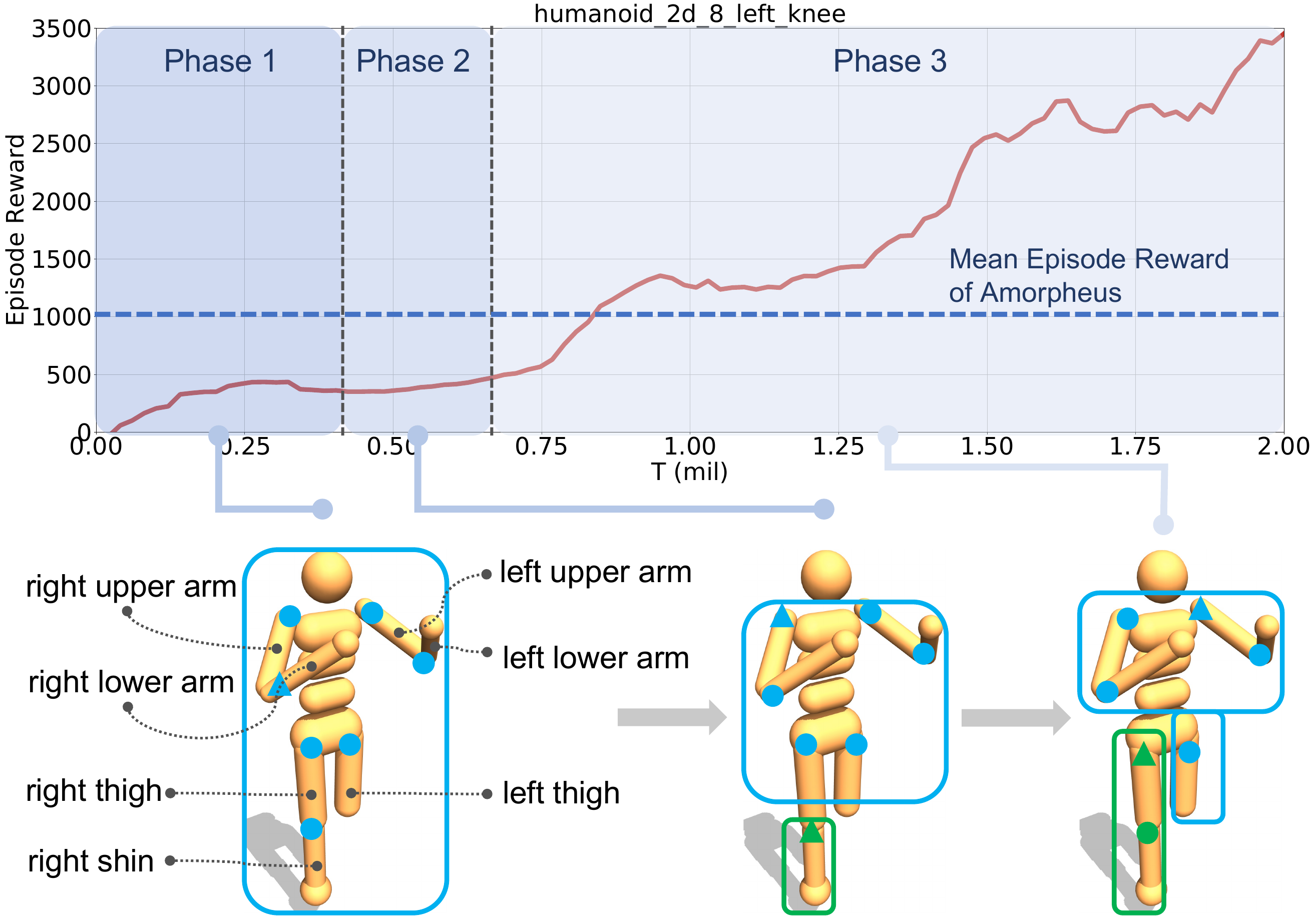}
\caption{Synergy structure evolution of \name{} in Humanoid. Phase are divided according to the change of synergy clusters and synergy clusters are masked with colored shapes.}\label{fig:evolution}
\end{figure}

To investigate why \name{} performs better than baselines, in this section, we visualize the synergy clusters learned by \name{} and the evolutionary process of synergy clusters in \Humanoids{}. 

Affinity propagation which we used as the method to discover synergy clusters can naturally outputs clustering results and the cluster centers in each cluster. We first visualize the synergy clusters of \name{} in \Humanoids{} at the end of training in Figure~\ref{fig:synergy_cluster}. Here we use different colors and two shapes to mark each joint in robots. Different colors represent different synergy clusters, and joints marked with the same color are in the same cluster. Joints marked with triangles are the centers of their corresponding clusters. For example, the \textit{Full humanoid} in the lowermost right corner has 4 synergy clusters, and the joints between the torso and two thighs and two arms are the cluster centers. 

To further investigate how these clusters evolve, we visualize the learning process of \textit{Left Knee Humanoid} in Figure~\ref{fig:evolution}. According to the change of synergy clusters, the learning process can be divided into three phases. In Phase 1, the episode reward is very low and $\Delta\tilde{Q}$ may only provide noisy information, which results in all joints being clustered in a same cluster. In Phase 2, \name{} finds that the right shin is important and divides it into a separate cluster. Then in Phase 3, \name{} finally obtains a suitable clustering solution and the episode reward increases correspondingly. 

The visualization results reflect two facts: (1) close joints are more likely to be in the same synergy cluster, and (2) joints near the torso may be more influential than those who are far from the torso, and are thus selected as the cluster centers. These clusters can reduce learning and controlling complexities but are hard to discover through a standard learning mechanism. Instead, \name{} makes use of early, less accurate learning information $\Delta\tilde{Q}$ and structural information via unsupervised learning to discover synergy clusters. The synergy clusters then further facilitate learning.

\subsection{Single-task with numerous joints}\label{sec:exp_singletask}
\begin{figure}[htbp]
\centering
\includegraphics[width=\linewidth]{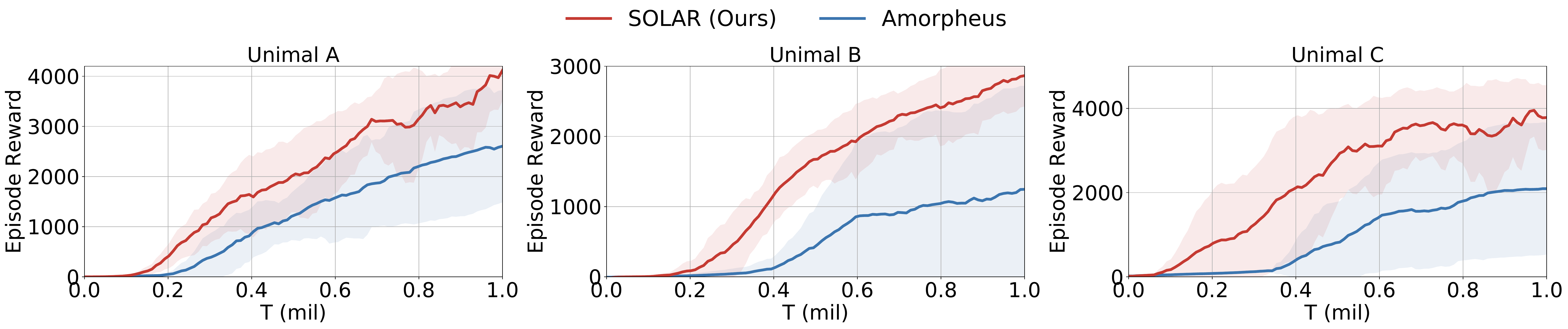}
\caption{Multi-Task performance of our method \name{} compared to \Amorpheus{}}\label{fig:STRL}
\end{figure}

To consolidate the speculation in Sec.~\ref{sec:exp_multitask} that \name{} is suitable for tasks with numerous joints, we test \name{} in three single-robot tasks sampled from UNIMALS~\citep{gupta2021embodied}. The average number of joints is about twice as many as that in \Humanoids{}. For more details about these tasks, please refer to Appendix~\ref{appx:repro}. We showcase the results in Figure~\ref{fig:STRL}. \name{} learns faster than state-of-the-art \Amorpheus{}, showing much higher sample efficiency, and again outperforms it in average episode rewards. 

\section{Conclusions}\label{sec:discussion}
In this paper, we use the inspiration of muscle synergies from neuroscience to reduce the control complexity of modular reinforcement learning in tasks with a large number of degree of freedom. We present \name{} that leverages both structural information and policy learning information to learn synergy clusters. Moreover, a stacked-transformer architecture is proposed for learning synergy actions which are combined linearly to produce low-rank actuator actions. The evaluations of \name{} on various tasks and settings showcase its effectiveness. The authors do not see obvious negative societal impacts of the proposed method.

\section*{Acknowledgments}
This work is supported in part by Science and Technology Innovation 2030 – “New Generation Artificial Intelligence” Major Project (No. 2018AAA0100904) and National Natural Science Foundation of China (62176135).

\bibliographystyle{unsrtnat}  
\bibliography{neurips_2022}  

\section*{Checklist}


\begin{enumerate}

\item For all authors...
\begin{enumerate}
  \item Do the main claims made in the abstract and introduction accurately reflect the paper's contributions and scope?
    \answerYes{}
  \item Did you describe the limitations of your work?
    \answerYes{} In Appendices.
  \item Did you discuss any potential negative societal impacts of your work?
    \answerYes{} 
  \item Have you read the ethics review guidelines and ensured that your paper conforms to them?
    \answerYes{}
\end{enumerate}

\item If you are including theoretical results...
\begin{enumerate}
  \item Did you state the full set of assumptions of all theoretical results?
    \answerNA{}
        \item Did you include complete proofs of all theoretical results?
    \answerNA{}
\end{enumerate}

\item If you ran experiments...
\begin{enumerate}
  \item Did you include the code, data, and instructions needed to reproduce the main experimental results (either in the supplemental material or as a URL)?
    \answerYes{} In Appendices and supplemental materials.
  \item Did you specify all the training details (e.g., data splits, hyperparameters, how they were chosen)?
    \answerYes{} In Appendices.
        \item Did you report error bars (e.g., with respect to the random seed after running experiments multiple times)?
    \answerYes{} All results have confidence intervals.
        \item Did you include the total amount of compute and the type of resources used (e.g., type of GPUs, internal cluster, or cloud provider)?
    \answerYes{} In Appendices.
\end{enumerate}

\item If you are using existing assets (e.g., code, data, models) or curating/releasing new assets...
\begin{enumerate}
  \item If your work uses existing assets, did you cite the creators?
    \answerYes{} In Sec.~\ref{sec:exp_setup}.
  \item Did you mention the license of the assets?
    \answerYes{} In Appendices.
  \item Did you include any new assets either in the supplemental material or as a URL?
    \answerYes{} Website for qualitative results.
  \item Did you discuss whether and how consent was obtained from people whose data you're using/curating?
    \answerNA{}
  \item Did you discuss whether the data you are using/curating contains personally identifiable information or offensive content?
    \answerNA{}
\end{enumerate}

\item If you used crowdsourcing or conducted research with human subjects...
\begin{enumerate}
  \item Did you include the full text of instructions given to participants and screenshots, if applicable?
    \answerNA{}
  \item Did you describe any potential participant risks, with links to Institutional Review Board (IRB) approvals, if applicable?
    \answerNA{}
  \item Did you include the estimated hourly wage paid to participants and the total amount spent on participant compensation?
    \answerNA{}
\end{enumerate}

\end{enumerate}


\newpage
\appendix

\section{Reproducibility}\label{appx:A}\label{appx:repro}

\subsection{Environment description}

We run experiments on two benchmarks as described in Table~\ref{tab:envs}. All these benchmarks are implemented in MuJoCo~\citep{todorov2012mujoco}. \citet{huang2020one} created \Walkers{}, \Humanoids{}, \Hoppers{} and \WHHs{}, by removing some limbs and joints from the original intact morphology of Gym MuJoCo robots~\citep{todorov2012mujoco,brockman2016openai}. In this benchmark, every robot has the ability to hop, walk, or run, \ie, only the robots that can move forward are left. Following \citet{huang2020one,kurin2020cage,hong2021structure}, we use this benchmark for the evaluation of multi-task learning (Sec.~\ref{sec:exp_multitask}) and zero-shot generalization ability (Sec.~\ref{sec:exp_zeroshot}). 
As for the single-task learning benchmark, we sampled some morphologies from UNIMALS~\citep{gupta2021embodied,gupta2021metamorph}. \citet{gupta2021embodied} created UNIMALS by evolution on robots' morphologies, kinematics, and dynamics. At the end of the evolution, they had 100 task optimized robots. We choose a subset of these robots that are optimized for locomotion in flat terrain environment for single-task evaluation in Sec.~\ref{sec:exp_singletask}. We ensure that the selected robots have different morphologies from each other. For multi-task evaluation, we use the whole 100 robots for training in Appendix~\ref{appx:MTRL_unimals}. And for zero-shot evaluation in robots' dynamic and kinematic properties in Appendix~\ref{appx:MTRL_unimals}, we use the test set created by \citet{gupta2021metamorph}, which contains 2400 robots. 

The reward of these tasks is given by the speed of the robot moving forward, and is penalized by the action norm. An episode terminates when the robot's height is low or the episode length exceeds 1000 time steps. For all the environments, we can extract the morphology of the robot as a graph from the corresponding MuJoCo XML file. Then the morphological information is concatenated in observations in baselines or is used to generate embeddings for action transformation in \name{}. The visualization of these robots can be found in synergy clustering results: \Walkers{} as in Figure~\ref{fig:synergy_cluster_walkers}, \Humanoids{} as in Figure~\ref{fig:synergy_cluster}, \Hoppers{} as in Figure~\ref{fig:synergy_cluster_hoppers} and \UNIMALS{} as in Figure~\ref{fig:synergy_cluster_unimals}. The number of actuators we evaluated in this work varies from three to nine.

\begin{table}[ht]
\centering
\caption{Full list of environments.}\label{tab:envs}
\begin{tabular}{@{}lll@{}}
\toprule
\textbf{Environment} & \textbf{Train set}                                                                                                                                                                                                & \textbf{Test set}                                                                                \\ \midrule
\Walkers{}              & \begin{tabular}[c]{@{}l@{}}walker\_2\_main\\ walker\_4\_main\\ walker\_5\_main\\ walker\_7\_main\end{tabular}                                                                                                     &                                                                                                  \\ \midrule
\Humanoids{}            & \begin{tabular}[c]{@{}l@{}}humanoid\_2d\_7\_left\_arm\\ humanoid\_2d\_lower\_arms\\ humanoid\_2d\_7\_right\_leg\\ humanoid\_2d\_8\_left\_knee\\ humanoid\_2d\_8\_right\_knee\\ humanoid\_2d\_9\_full\end{tabular} & \begin{tabular}[c]{@{}l@{}}humanoid\_2d\_7\_left\_leg\\ humanoid\_2d\_7\_right\_arm\end{tabular} \\ \midrule
\Hoppers{}              & \begin{tabular}[c]{@{}l@{}}hopper\_3\\ hopper\_4\\ hopper\_5\end{tabular}                                                                                                                                         &                                                                                                  \\ \midrule
\multicolumn{3}{l}{\WHHs{}. Union of \Walkers{}, \Humanoids{} and \Hoppers{}} \\ \midrule
UNIMALS              & unimalA, unimalB, unimalC                                                                                                                                                                                         &                                                                                                  \\ \bottomrule
\end{tabular}
\end{table}

\subsection{Implementation details}\label{appx:imple}

We implement \name{} based on \Amorpheus{}~\citep{kurin2020cage}, which is built on Official PyTorch Tutorial~\citep{Seq2Seq}. \Amorpheus{} also shares the codebase with \SMP{}~\citep{huang2020one}. Table~\ref{tab:hypers} provides the hyperparameters needed to replicate our experiments. 

\begin{table}[h]
\centering
\caption{Hyperparameters of our \name{}.}\label{tab:hypers}
\begin{tabular}{@{}cc@{}}
\toprule
\textbf{Hyperparameter}        & \textbf{Value} \\ \midrule
Learning rate                  & 0.0001         \\
Gradient clipping              & 0.1            \\
Normalization                  & LayerNorm      \\
Total attention layers         & 3              \\
Intra-synergy attention layers & 1              \\
Inter-synergy attention layers & 2              \\
Attention heads                & 2              \\
Attention hidden size          & 256            \\
Encoder output size            & 128            \\
Mini-batch size                & 100            \\
Replay buffer size             & 10M            \\
Attention embedding size        & 128            \\ \bottomrule
\end{tabular}
\end{table}

As in Figure~\ref{fig:framework}, there are three main components in our implementation, \ie, \textit{Intra-Synergy Self-Attention}, \textit{Inter-Synergy Self-Attention} and \textit{Action Transformation}. To achieve \textit{Intra-Synergy Self-Attention} efficiently, we pass a synergy-based mask into PyTorch's \textit{TranformerEncoderLayer} through \textit{src\_mask} function argument. The synergy-based mask is discussed in detail in Sec.~\ref{sec:synergy_policy}. \textit{Inter-Synergy Self-Attention} is implemented by the normal \textit{TranformerEncoderLayer} without masks. 

As for \textit{Action Transformation}, the main challenge is how to obtain a transformation matrix $\mathbf{T}\in\mathbb{R}^{K_n\times L_n}$, where $K_n$ and $L_n$ are the number of actuators and the number of synergies of robot $n\in\{1,2,\cdots,N\}$. $L_n$ may change during the learning process. Intuitively, the matrix $\mathbf{T}$ is dependent on the robot's morphology. Here we use a traversal-based positional embedding~\citep{hong2021structure} technique, and these embeddings together represent the robot's morphological information. To obtain the traversal-based positional embedding, we first apply left-child-right-sibling representation to represent a general tree as a binary tree. Then, we traverse the binary tree in pre-order, in-order and post-order, which forms a triple for each actuator consisting of its orders. And these triples together are sufficient to reconstruct the original tree, which indicates that each triple contains the structural information of the corresponding actuator. The triple of each joint serves as indices into a embedding pool $\{\vr_1, \vr_2, \dots, \vr_p\}$ where $p=\max_n K_n$ is the maximum number of actuators in a robot, and each $\vr_i\in\R^s$ is a learnable vector. The selected embeddings are processed by a network, and we concatenate the outputs to obtain a representation of the corresponding actuator. Using dot product between representations of two actuators, we get a matrix $\mathbf{H}\in\mathbb{R}^{K_n\times K_n}$ where $\mathbf{H}_{i,j}$ indicates the relation between actuator $i$ and actuator $j$. Finally, we average the columns corresponding to actuators in the same synergy cluster and obtain the transformation matrix $\mathbf{T}\in\mathbb{R}^{K_n\times L_n}$. The embedding pool and network are updated by backpropagating the RL loss.

Experiments are carried out on NVIDIA GTX 2080 Ti GPUs. For \Humanoids{}, our method requires approximately 10G of RAM and 5G of video memory, and takes about 23 hours to finish 2M timesteps of training.

The code for our method is included in the supplementary materials with sufficient setup and running instructions. The code follows the MIT license.

\subsection{Baselines and ablations}

We compare our method against various baselines and ablations. Here we explain the implementations of these baselines and ablations.

\Amorpheus{}. We use the original implementation of \Amorpheus{} released by \citet{kurin2020cage}. For fair comparison, \name{} uses the same value for those hyperparameters that are shared with \Amorpheus{} (Table~\ref{tab:hypers}). 

\SMP{}. We use the implementation of \SMP{} in the \Amorpheus{} codebase, which is the same as the original implementation of \SMP{} provided by \citet{huang2020one}.

\Monolithic{}. Following the setup by \citet{huang2020one}, we choose TD3 as the standard monolithic RL baseline.  The actor and critics of TD3 are implemented by fully-connected neural networks, which take the concatenation of observations of all actuators as input. Since number of actuators varies in different robots, the dimension of observations is incompatible. To overcome this issue, we zero-pad the observations and actions to the maximum dimension across all robots. 

\Amorpheuswmask{}. We passed a synergy-based mask to the first layer of self-attention in \Amorpheus{}, and the mask is learned in the same way as in our method. The other two layers of \Amorpheus{} are not modified.  

\namewopreference{}. We simply set the preference vector of the affinity propagation clustering algorithm to None, and the algorithm will use the median of the input affinity matrix as a default preference vector.

\section{Additional experiments and discussions}\label{appx:add_exp}

\subsection{\UNIMALS{} \Locomotion{} task and \Manipulation{} task}\label{appx:MTRL_unimals}


\begin{figure}[htbp]
\centering
\includegraphics[width=\linewidth]{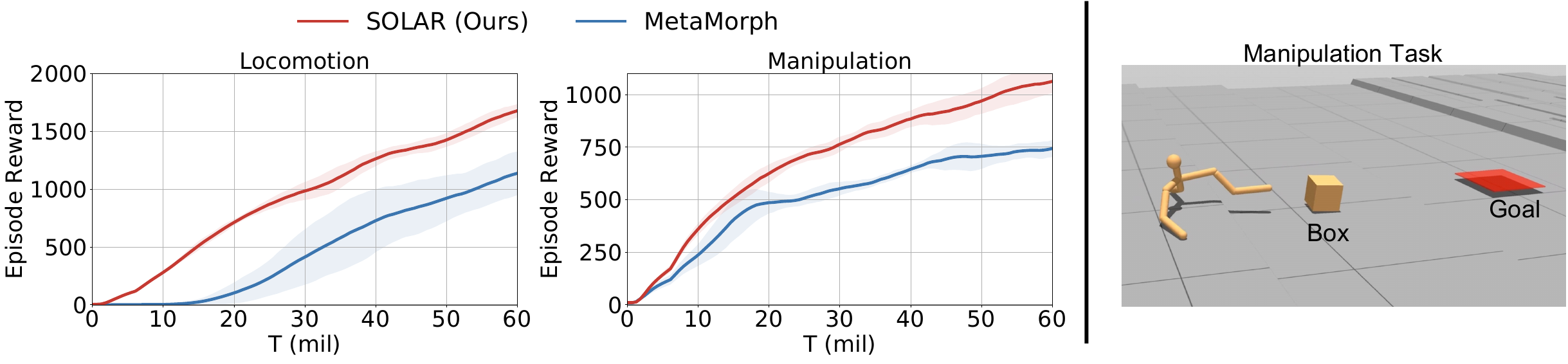}
\caption{Left: Multi-Task performance in \UNIMALS{} of our method \name{} compared to \MetaMorph{}. Right: \Manipulation{} task.}\label{fig:UNIMALS_train}
\end{figure}

To further analyze the scalability of our \name{} when the number of robots and the number of actuators are large, we use the original UNIMALS~\citep{gupta2021embodied} \Locomotion{} tasks in flat terrain and \Manipulation{} tasks in variable terrain. In \Locomotion{} tasks, the robot needs to go forward as fast as possible. In \Manipulation{} tasks, the robot needs to first reach a box, and then push this box to a randomly generated goal. The UNIMALS has 100 robots with different morphologies, kinematics, and dynamics for multi-task training. We compared our method \name{} against \MetaMorph{}~\citep{gupta2021metamorph}. In this subsection, \name{} is built upon \MetaMorph{}. Like \MetaMorph{}, \name{} incorporates structural information into \Amorpheus{} and uses a dynamic replay buffer balancing technique to deal with the large number of robots. \name{} can be simply combined with \MetaMorph{} by modifying its self-attention structure. The training results are shown in Figure~\ref{fig:UNIMALS_train}. \name{} outperforms \MetaMorph{} by a large margin, which suggests that \name{} is also effective with numerous different robots.

\begin{table}[h]
\centering
\caption{Zero-shot generalization performance of \name{} compared against \MetaMorph{} in \UNIMALS{} \Locomotion{} task.} \label{tab:UNIMALS_zeroshot}
\begin{tabular}{@{}c|l|ll@{}}
\toprule
\multicolumn{1}{l|}{}                            & Variants      & \name{} & \MetaMorph{} \\ \midrule
\multirow{4}{*}{Dynamics}                        & Armature      & \textbf{1253.39} $\pm$ 28.25    & 843.31 $\pm$ 14.80         \\
                                                 & Density       & \textbf{1654.43} $\pm$ 24.55    & 1089.64 $\pm$ 22.64        \\
                                                 & Damping       & \textbf{1663.64} $\pm$ 26.21    & 1113.04 $\pm$ 10.18         \\
                                                 & Gear          & \textbf{1480.45} $\pm$ 30.49    & 987.29 $\pm$ 8.05         \\ \midrule
\multicolumn{1}{l|}{\multirow{2}{*}{Kinematics}} & Module param. & \textbf{1084.41} $\pm$ 13.93    & 700.43 $\pm$ 17.02        \\
\multicolumn{1}{l|}{}                            & Joint angle   & \textbf{477.50} $\pm$ 6.83    & 274.37 $\pm$ 6.55         \\ \bottomrule
\end{tabular}
\end{table}

\begin{table}[h]
\centering
\caption{Zero-shot generalization performance of \name{} compared against \MetaMorph{} in \UNIMALS{} \Manipulation{} task.}\label{tab:UNIMALS_zeroshot_mani}
\begin{tabular}{@{}c|l|ll@{}}
\toprule
\multicolumn{1}{l|}{}                            & Variants      & \name{} & \MetaMorph{} \\ \midrule
\multirow{4}{*}{Dynamics}                        & Armature      & \textbf{792.69} $\pm$ 32.35    & 685.36 $\pm$ 9.03         \\
                                                 & Density       & \textbf{943.07} $\pm$ 57.46    & 823.20 $\pm$ 18.95        \\
                                                 & Damping       & \textbf{984.19} $\pm$ 24.14    & 860.70 $\pm$ 11.24         \\
                                                 & Gear          & \textbf{862.04} $\pm$ 36.90    & 784.54 $\pm$ 12.01         \\ \midrule
\multicolumn{1}{l|}{\multirow{2}{*}{Kinematics}} & Module param. & \textbf{663.18} $\pm$ 12.47    & 611.43 $\pm$ 5.80        \\
\multicolumn{1}{l|}{}                            & Joint angle   & 314.16 $\pm$ 4.95    & \textbf{455.49} $\pm$ 11.28         \\ \bottomrule
\end{tabular}
\end{table}

In Sec.~\ref{sec:exp_zeroshot}, we test the zero-shot performance of \name{} on robots with different morphologies. However, in reality, there are cases where the morphologies are not changed, but the dynamics (armature, density, damping, and motor gear) and kinematics (module shape parameters and joint angles) of robots are new in unseen tasks. Thus we also benchmark the zero-shot performance of \name{} in these cases. \citet{gupta2021metamorph} created a test set to test zero-shot performance based on the 100 training robots of \UNIMALS{}. For each robot in the training set, they create 4 different variants for each property (regarding the dynamics and kinematics), leading to a total number of $100\times 6\times 4 = 2400$ test variants. Here we reproduce from their paper the sampling ranges of test variants in Table~\ref{tab:UNIMALS_variants} . Please refer to \citet{gupta2021metamorph} for more details. Due to the limitation of computing resources, we reduce the batch size of \name{} and \MetaMorph{} from 5120 to 128, and other hyperparameters follow \citet{gupta2021metamorph}.

To evaluate the zero-shot generalization performance, we load our models trained on multi-task UNIMALS and show the averaged performance (and variance) over 4 random seeds in Table~\ref{tab:UNIMALS_zeroshot} and Table~\ref{tab:UNIMALS_zeroshot_mani} for \Locomotion{} task and \Manipulation{} task, respectively. We find that \name{} exhibits strong generalization performance compared to the baseline, which indicates that the synergy clusters and action transformations learned on the training set are robust to dynamic and kinematic variations.

\begin{table}[]
\centering
\caption{Dynamics and kinematics variations to generate the test robot set, reproduced from \citet{gupta2021metamorph}.}\label{tab:UNIMALS_variants}
\begin{tabular}{@{}lr@{}}
\toprule
\multicolumn{2}{c}{\textbf{Kinematics}}                                                                                                                                                                       \\ \midrule
\textbf{Variation type} & \textbf{Value}                                                                                                                                                                      \\ \midrule
Limb radius             & {[}0.03, 0.05{]}                                                                                                                                                                    \\
Limb height             & {[}0.15, 0.45{]}                                                                                                                                                                    \\
Joint angles            & \begin{tabular}[c]{@{}r@{}}{[}(-30, 0), (0, 30), (-30, 30),\\ (-45, 45), (-45, 0), (0, 45),\\ (-60, 0), (0, 60), (-60, 60),\\ (-90, 0), (0, 90), (-60, 30)(-30, 60){]}\end{tabular} \\ \midrule
\multicolumn{2}{c}{\textbf{Dynamics}}                                                                                                                                                                         \\ \midrule
Armature                & {[}0.1, 2{]}                                                                                                                                                                        \\
Density                 & {[}0.8, 1.2{]} $\times$ limb density                                                                                                                                                       \\
Damping                 & {[}0.01, 5.0{]}                                                                                                                                                                     \\
Gear                    & {[}0.8, 1.2{]} $\times$ motor gear                                                                                                                                                         \\ \bottomrule
\end{tabular}
\end{table}

\subsection{Compared with other baselines in single-task}

\begin{figure}[htbp]
\centering
\includegraphics[width=\linewidth]{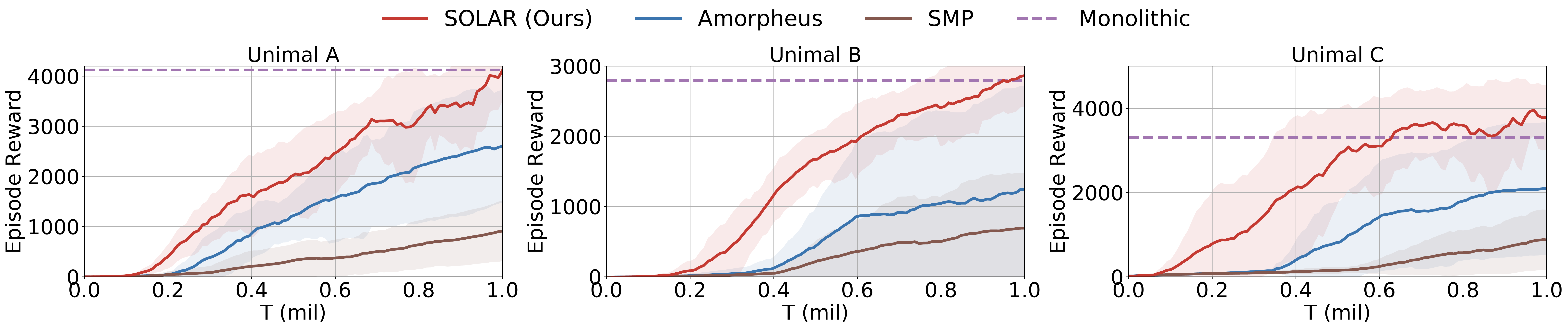}
\caption{Multi-Task performance of our method \name{} compared to \Amorpheus{}, \SMP{} and the final performance of \Monolithic{}.}\label{fig:STRL_more}
\end{figure}

In Sec.~\ref{sec:exp_singletask}, we compare \name{} with transformer-based baseline \Amorpheus{}, here we also compare our \name{} with the GNN-style baseline \SMP{} in single-task evaluations in Figure~\ref{fig:STRL_more} as additional results.

\subsection{Synergy clustering results of more morphologies}

\begin{figure}[htbp]
\centering
\includegraphics[width=0.65\linewidth]{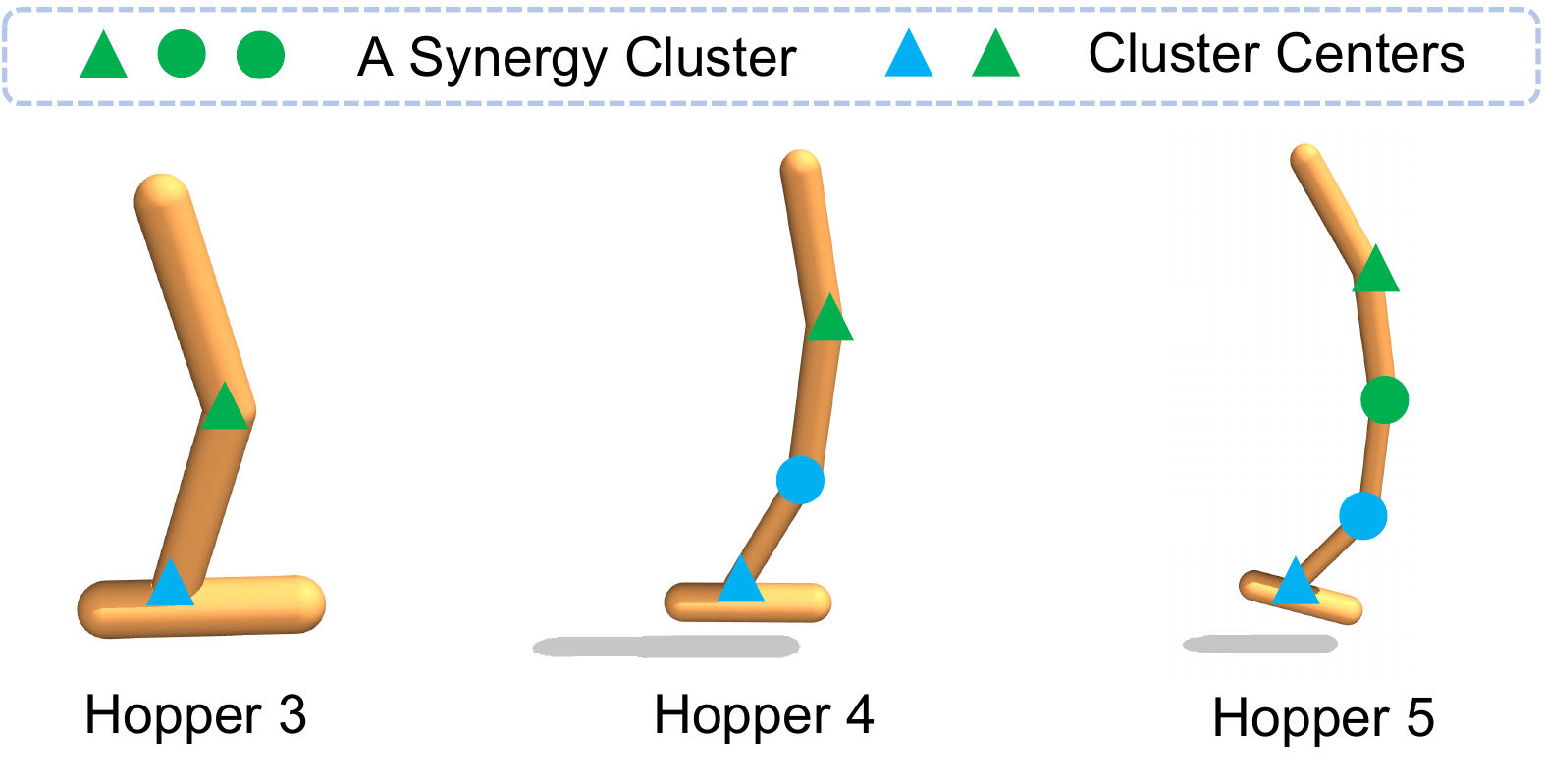}
\caption{Synergy clustering results of \name{} in \Hoppers{}. Different colors represent different synergy clusters, and joints marked with the same color are in the same cluster. Joints marked with triangles are the centers of their corresponding clusters.}\label{fig:synergy_cluster_hoppers}
\centering
\includegraphics[width=0.9\linewidth]{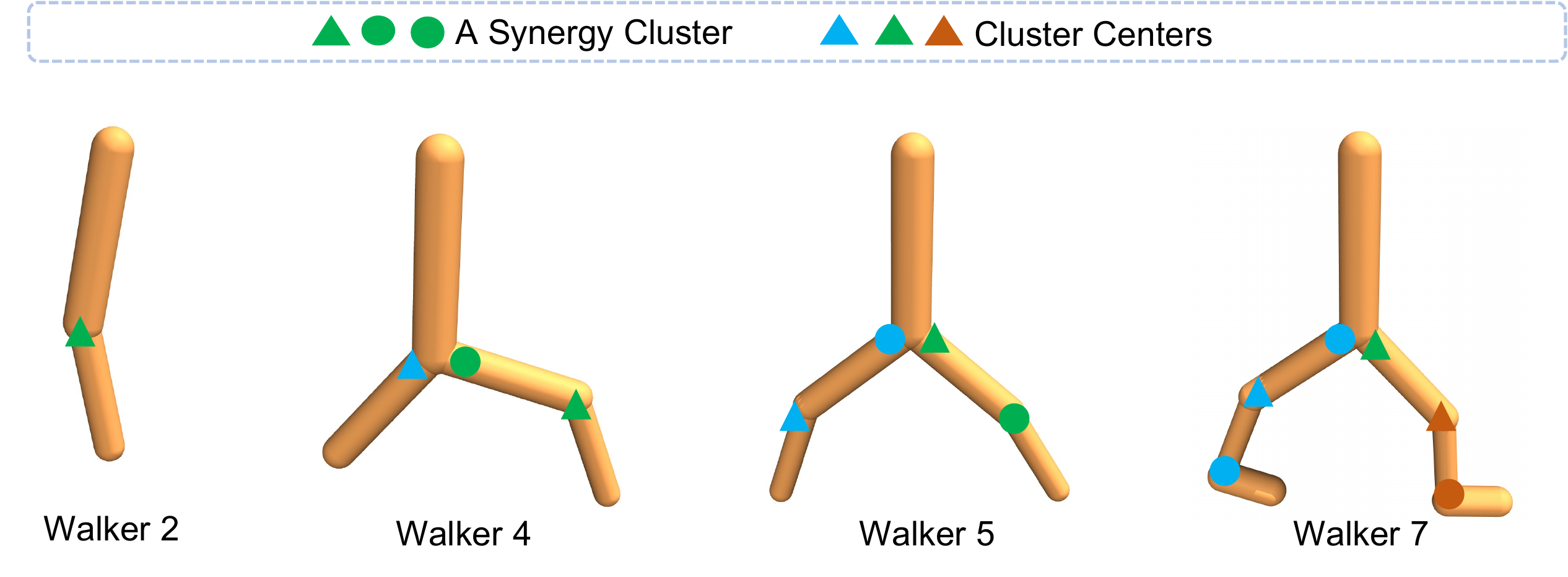}
\caption{Synergy clustering results of \name{} in \Walkers{}.}\label{fig:synergy_cluster_walkers}
\centering
\includegraphics[width=0.8\linewidth]{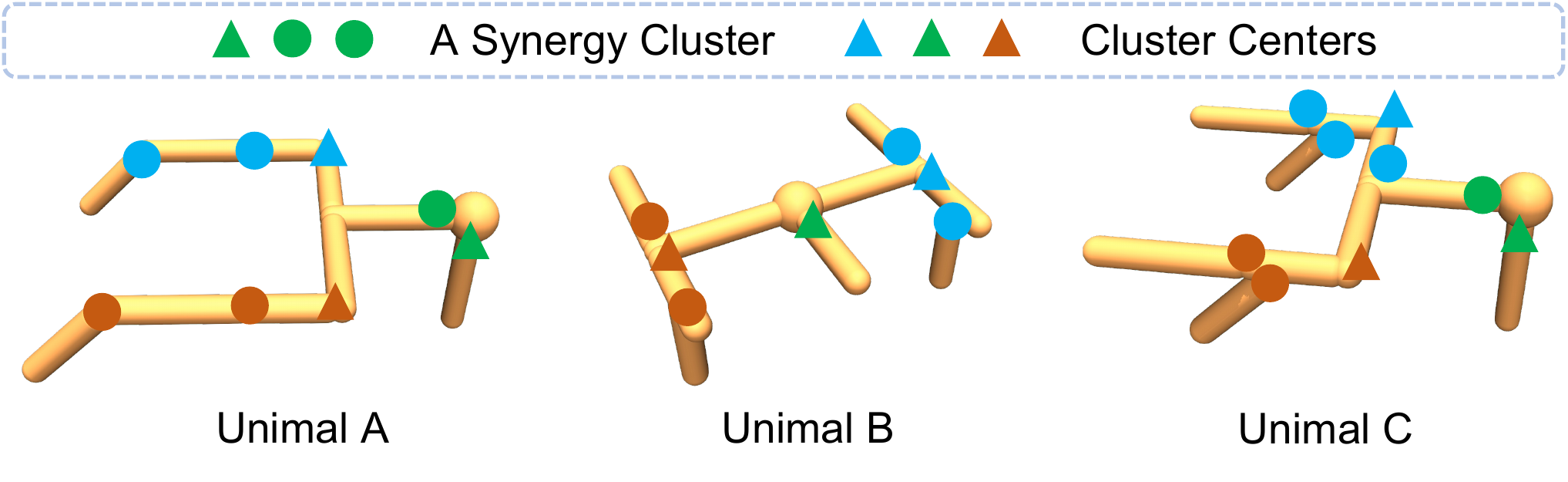}
\caption{Synergy clustering results of \name{} in \UNIMALS{}.}\label{fig:synergy_cluster_unimals}
\end{figure}

In Figure~\ref{fig:synergy_cluster}, we visualize the synergy clustering results of \Humanoids{} to analyze the effect of synergy. In this section, we provide visualizations of synergy clustering results for \Hoppers{} (Figure~\ref{fig:synergy_cluster_hoppers}), \Walkers{} (Figure~\ref{fig:synergy_cluster_walkers}), and \UNIMALS{} (Figure~\ref{fig:synergy_cluster_unimals}) as additional results. 

These additional results further consolidate our analysis in Sec.~\ref{sec:exp_analysis}: (1) close joints are more likely to be in the same synergy cluster, and (2) joints near the torso may be more influential than those who are far from the torso, and are thus selected as the cluster centers.

\subsection{Connection to conventional notion of muscle synergy}
Mathematically, conventional muscle synergy studies typically use non-negative matrix factorization (Rabbi et al, 2020) and aim to solve the following optimization problem:
$$
\min_{H\in\mathbb{R}_{+}^{N\times M},A\in\mathbb{R}_{+}^{M\times T}} \|U - HA \|_F,
$$
where $U\in \mathbb{R}^{N\times T}$ is a given matrix of observed electrical control signals. The element at $i$th row and $t$th column of $U$ is the control signal to muscle $i$ at timestep $t$. In this optimization problem, the number of synergies, $M$ (where $M<N$), is typically pre-defined or chosen according to a pre-defined reconstruction error threshold. By solving this problem, one can discover the synergy structure by observing matrix $H$. Conventional muscle synergy studies that use other factorization methods (PCA, ICA, and FA) share the similar optimization problem with different matrix constraints. 

We study a similar optimization problem but with additional constraints:
$$
\max_{U,H,A} \sum_t \gamma^t R(s_t, U_t) - \|U - HA \|_F.
$$
Here $s_t$ is the environment state at timestep $t$, and $U_t$ is the column $t$ of matrix $U$, i.e., actions at timestep $t$. And $R(s_t,U_t)$ is the reward of choosing action $U_t$ at state $s_t$.

The differences are:

\begin{enumerate}
    \item We additionally maximize the expected return of muscle actions. 
    \item In our formulation, the number of synergies is not pre-defined but is learned in an unsupervised manner.
    \item The matrix $U$ is also not given, but is generate by an attention-based policy. This policy are optimized to maximize return as well as to minimize the decomposition loss (term 2). 
\end{enumerate}

In summary, we share a common optimization object with conventional muscle synergy studies, which is a decomposition loss. However, we need to additionally learn the number of synergies and control signals which are inputs in the conventional studies. Structurally, our framework gives an attention function class that covers a synergy decomposition solution which can minimize the decomposition loss while enable an efficient control policy.

\subsection{Limitations and future works}


When generalizing to unseen robots with larger numbers of actuators than training robots, the embeddings of testing robots are not learned, which will hamper the zero-shot performance. One possible future direction will involve designing a more scalable actuator embedding method. Moreover, SOLAR is more suitable for robots with large number of actuators. Our approach is able to reduce a great degree of control complexity for these robots. But for robots with few actuators, SOLAR may only have a little positive impact and the learning of synergy-aware policy may even damage the performance.

\end{document}